\documentclass[namedreferences]{article}
\usepackage{theapa}
\usepackage{epsfig} 
\usepackage{psfig}
\usepackage{xspace}
\usepackage{url}

\usepackage{latexsym}           
\usepackage{endnotes}           
     
\newcommand{\POS}{\texttt{\bf p}}
\newcommand{\NEG}{\texttt{\bf n}}
\newcommand{\YES}{\texttt{\bf Y}}
\newcommand{\NO}{\texttt{\bf N}}

\newcommand{\rocch}{\textsc{rocch}}

\newcommand{\IF}{\textbf{if~}}
\newcommand{\THEN}{\textbf{then~}}
\newcommand{\ELSE}{\textbf{else~}}
\newcommand{\ENDIF}{\textbf{end if}}

\newcommand{\ENDWHILE}{\textbf{end while}}

\newcommand{\WHILE}{\textbf{while~}}
\newcommand{\DO}{\textbf{do~}}

\newcommand{\EndProof}{$\Box$}

\newtheorem{theorem}{Theorem}
\newtheorem{lemma}[theorem]{Lemma}
\newtheorem{corollary}[theorem]{Corollary}
\newtheorem{definition}{Definition}

\newcommand{\Partial}[2]{\frac{\partial #1}{\partial #2}}

\newcommand{\mlc}{\ensuremath{\mathcal{MLC\hspace{-.05em}\raisebox{.4ex}{\tiny\bf ++}}}}

\graphicspath{
  {./}
  {./Figs/}
  }

\newcommand{\eg}{{e.g.},\xspace} 
\newcommand{\ie}{{i.e.},\xspace}

\begin{document}

\centerline{\textbf{\Large Robust Classification for Imprecise Environments}}
\vspace{1ex}

\begin{flushleft}
  Foster Provost \hfill \texttt{provost@acm.org}\\
  \hspace*{.1in}\textit{New York University, New York, NY 10012}\\
  Tom Fawcett \hfill \texttt{tfawcett@acm.org}\\
  \hspace*{.1in}\textit{Hewlett-Packard Laboratories, Palo Alto, CA 94304}\\
  \vspace*{.2in}
\end{flushleft}

\begin{abstract}
  In real-world environments it usually is difficult to specify target
  operating conditions precisely, for example, target misclassification costs.
  This uncertainty makes building robust classification systems problematic.
  We show that it is possible to build a hybrid classifier that will perform
  at least as well as the best available classifier for any target conditions.
  In some cases, the performance of the hybrid actually can surpass that of
  the best known classifier.  This robust performance extends across a wide
  variety of comparison frameworks, including the optimization of metrics such
  as accuracy, expected cost, lift, precision, recall, and workforce
  utilization.  The hybrid also is efficient to build, to store, and to
  update.  The hybrid is based on a method for the comparison of classifier
  performance that is robust to imprecise class distributions and
  misclassification costs.  The ROC convex hull (\rocch) method combines
  techniques from ROC analysis, decision analysis and computational geometry,
  and adapts them to the particulars of analyzing learned classifiers.  The
  method is efficient and incremental, minimizes the management of classifier
  performance data, and allows for clear visual comparisons and sensitivity
  analyses.  Finally, we point to empirical evidence that a robust hybrid
  classifier indeed is needed for many real-world problems.
\end{abstract}  

\begin{flushleft}
  \textbf{Keywords:} classification, learning, uncertainty, evaluation,
  comparison, multiple models, cost-sensitive learning, skewed distributions\\

  \vspace*{.1in}
  \textbf{\large To appear in \emph{Machine Learning Journal}}

\end{flushleft}

\vspace{.1in}

\section{Introduction}

Traditionally, classification systems have been built by experimenting with
many different classifiers, comparing their performance and choosing the best.
Experimenting with different induction algorithms, parameter settings, and
training regimes yields a large number of classifiers to be evaluated and
compared.  Unfortunately, comparison often is difficult in real-world
environments because key parameters of the target environment are not known.
The optimal cost/benefit tradeoffs and the target class priors seldom are
known precisely, and often are subject to change
\cite{ZahaviLevin:1997:issues_probl_applying_neural_comput,FriedmanWyatt:97,KlinkenbergJoachims:2000}.
For example, in fraud detection we cannot ignore misclassification costs or
the skewed class distribution, nor can we assume that our estimates are
precise or static \cite{FawcettProvost:97}.  We need a method for the
management, comparison, and application of multiple classifiers that is robust
in imprecise and changing environments.

We describe the \textit{ROC convex hull} (\rocch) method, which combines
techniques from ROC analysis, decision analysis and computational geometry.
The ROC convex hull decouples classifier performance from specific class and
cost distributions, and may be used to specify the subset of methods that are
potentially optimal under any combination of cost assumptions and class
distribution assumptions.  The \rocch\ method is efficient, so it facilitates
the comparison of a large number of classifiers.  It minimizes the management
of classifier performance data because it can specify exactly those
classifiers that are potentially optimal, and it is incremental, easily
incorporating new and varied classifiers without having to reevaluate all
prior classifiers.

We demonstrate that it is possible and desirable to avoid complete commitment
to a single best classifier during system construction.  Instead, the \rocch\ 
can be used to build from the available classifiers a hybrid classification
system that will perform best under any target cost/benefit and class
distributions.  Target conditions can then be specified at run time.
Moreover, in cases where precise information is still unavailable when the
system is run (or if the conditions change dynamically during operation), the
hybrid system can be tuned easily (and optimally) based on feedback from its
actual performance.

The paper is structured as follows.  First we sketch briefly the traditional
approach to building such systems, in order to demonstrate that it is brittle
under the types of imprecision common in real-world problems.  We then
introduce and describe the \rocch\ and its properties for comparing and
visualizing classifier performance in imprecise environments.  In the
following sections we formalize the notion of a robust classification system,
and show that the \rocch\ is an elegant method for constructing one
automatically.  The solution is elegant because the resulting hybrid
classifier is robust for a wide variety of problem formulations, including the
optimization of metrics such as accuracy, expected cost, lift, precision,
recall, and workforce utilization, and it is efficient to build, to store, and
to update.  We then show that the hybrid actually can do better than the best
known classifier in certain situations.  Finally, by citing results from
empirical studies, we provide evidence that this type of system indeed is
needed.

\subsection{An example}

A systems-building team wants to create a system that will take a
large number of instances and identify those for which an action
should be taken.  The instances could be potential cases of fraudulent
account behavior, of faulty equipment, of responsive customers, of
interesting science, etc.  We consider problems for which the best
method for classifying or ranking instances is not well defined, so
the system builders may consider machine learning methods, neural
networks, case-based systems, and hand-crafted knowledge bases as
potential classification models.  Ignoring for the moment issues of
efficiency, the foremost question facing the system builders is: which
of the available models performs ``best'' at classification?

Traditionally, an experimental approach has been taken to answer this question,
because the distribution of instances can be sampled if it is not known a
priori.  The standard approach is to estimate the error rate of each model
statistically and then to choose the model with the lowest error rate.  This
strategy is common in machine learning, pattern recognition, data mining,
expert systems and medical diagnosis.  In some cases, other measures such as
cost or benefit are used as well.  Applied statistics provides methods such as
cross-validation and the bootstrap for estimating model error rates and recent
studies have compared the effectiveness of different methods
\cite{Dietterich:98,kohavi-accest,Salzberg:97}.

Unfortunately, this experimental approach is brittle under two types
of imprecision that are common in real-world environments.
Specifically, costs and benefits usually are not known precisely, and
target (prior) class distributions often are known only approximately
as well.  This observation has been made by many authors
\cite{Bradley:97,Catlett:95,ProvostFawcett:97}, and is in fact the
concern of a large subfield of decision analysis
\cite{WeinsteinFineberg:80}.  Imprecision also arises because the
environment may change between the time the system is conceived and
the time it is used, and even as it is used.  For example, levels of
fraud and levels of customer responsiveness change continually over
time and from place to place.

\subsection{Basic terminology}

\begin{figure}[tb]
  \begin{center}
    \epsfig{file=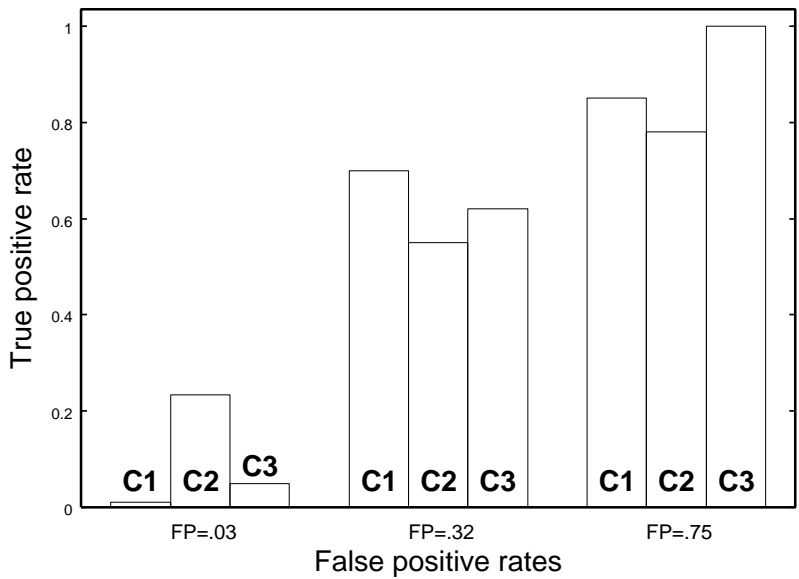,height=3in}
    \caption{Three classifiers under three different Neyman-Pearson decision
      criteria} 
    \label{fig:NP}
  \end{center}
\end{figure}

In this paper we address two-class problems.  Formally, each instance
$I$ is mapped to one element of the set $\{\POS,\NEG\}$ of (correct)
positive and negative classes.  A \emph{classification model} (or
\emph{classifier}) is a mapping from instances to predicted classes.
Some classification models produce a continuous output (\eg an
estimate of an instance's class membership probability) to which
different thresholds may be applied to predict class membership.  To
distinguish between the actual class and the predicted class of an
instance, we will use the labels $\{\YES,\NO\}$ for the
classifications produced by a model.  For our discussion, let
$c(\textit{classification}, \textit{class})$ be a two-place error cost
function where $c(\YES,\NEG)$ is the cost of a false positive error
and $c(\NO,\POS)$ is the cost of a false negative error.\footnote{For
this paper, we consider error costs to include benefits not realized,
and ignore the costs of correct classifications.}
We represent class distributions by the classes' prior probabilities
$p(\POS)$ and $p(\NEG) = 1 - p(\POS)$.

The true positive rate, or hit rate, of a classifier is:
\begin{displaymath}
  TP = p(\YES|\POS) \approx \frac{\rm positives\: correctly\: classified}
                   {\rm total\: positives}
\end{displaymath}
The false positive rate, or false alarm rate, of a classifier is:
\begin{displaymath}
  FP = p(\YES|\NEG) \approx \frac{\rm negatives\: incorrectly\: classified}
                   {\rm total\: negatives}
\end{displaymath}

The traditional experimental approach is brittle because it chooses
one model as ``best'' with respect to a specific set of cost functions
and class distribution.  If the target conditions change, this system
may no longer perform optimally, or even acceptably.  As an example,
assume that we have a maximum false positive rate $FP$, that must not
be exceeded.  We want to find the classifier with the highest possible
true positive rate, $TP$, that does not exceed the $FP$ limit.  This
is the Neyman-Pearson decision criterion \cite{Egan:75}.  Three
classifiers, under three such $FP$ limits, are shown in
figure~\ref{fig:NP}.  A different classifier is best for each $FP$
limit; any system built with a single ``best'' classifier is brittle
if the $FP$ requirement can change.

\section{Evaluating and visualizing classifier performance}

\subsection{Classifier comparison: decision analysis and ROC analysis}

Most prior work on building classifiers uses classification accuracy (or,
equivalently, undifferentiated error rate) as the primary evaluation metric.
The use of accuracy assumes that the class priors in the target environment
will be \textit{constant and relatively balanced}.  In the real world this
rarely is the case.  Classifiers often are used to sift through a large
population of normal or uninteresting entities in order to find a relatively
small number of unusual ones; for example, looking for defrauded accounts
among a large population of customers, screening medical tests for rare
diseases, and checking an assembly line for defective parts.  Because the
unusual or interesting class is rare among the general population, the class
distribution is very skewed
\cite{EzawaEtal:96,FawcettProvost:96,FawcettProvost:97,KubatHolteMatwin:98,SaittaNeri:98}.

As the class distribution becomes more skewed, evaluation based on accuracy
breaks down.  Consider a domain where the classes appear in a 999:1 ratio.  A
simple rule---always classify as the maximum likelihood class---gives a 99.9\%
accuracy.  This accuracy may be quite difficult for an induction algorithm
to beat, though the simple rule presumably is unacceptable if a non-trivial
solution is sought.  Skews of $10^2$ are common in fraud detection and skews
exceeding $10^6$ have been reported in other applications
\cite{ClearwaterStern:91}.

Evaluation by classification accuracy also assumes \textit{equal error costs}:
$c(\YES,\NEG)=c(\NO,\POS)$.  In the real world classifications lead to
actions, which have consequences.  Actions can be as diverse as denying a
credit charge, discarding a manufactured part, moving a control surface on an
airplane, or informing a patient of a cancer diagnosis.  The consequences may
be grave, and performing an incorrect action may be very costly.  Rarely are
the costs of mistakes equivalent.  In mushroom classification, for example,
judging a poisonous mushroom to be edible is far worse than judging an edible
mushroom to be poisonous.  Indeed, it is hard to imagine a domain in which a
classification system may be indifferent to whether it makes a false positive
or a false negative error.  In such cases, accuracy maximization should be
replaced with cost minimization.

The problems of unequal error costs and uneven class distributions are
related.  It has been suggested that, for training, high-cost
instances can be compensated for by increasing their prevalence in an
instance set \cite{bre84}.  Unfortunately, little work has been
published on either problem.  There exist several dozen articles in
which techniques for cost-sensitive learning are suggested
\cite{Turney-cost-bib}, but few studies evaluate and compare them
\cite{Domingos:99,pazzani-cost:94,ProvostFawcettKohavi:98}.  The
literature provides even less guidance in situations where
distributions are imprecise or can change.

\begin{figure}[tb]
  \begin{center}
    \epsfig{file=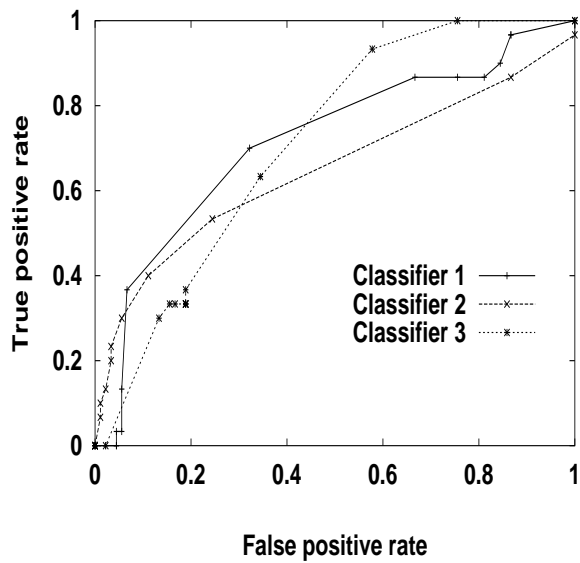,height=3in,width=3.2in}
    \caption{ROC graph of three classifiers}
    \label{fig:ROC-curves}
  \end{center}
\end{figure}

Given an estimate of $p(\POS|I)$, the posterior probability of an instance's
class membership, decision analysis gives us a way to produce cost-sensitive
classifications \cite{WeinsteinFineberg:80}.  Classifier error frequencies can
be used to approximate such probabilities \cite{pazzani-cost:94}.  For an
instance $I$, the decision to emit a positive classification from a particular
classifier is:

\[
[1-p(\POS|I)] \cdot c(\YES,\NEG) \; < \; p(\POS|I) \cdot c(\NO,\POS)
\]

Regardless of whether a classifier produces probabilistic or binary
classifications, its normalized cost on a test set can be evaluated 
empirically as:
\[
\textrm{Cost} = FP\cdot c(\YES,\NEG) + (1 - TP)\cdot c(\NO,\POS)
\]
Most published work on cost-sensitive classification uses an equation such as
this to rank classifiers.  Given a set of classifiers, a set of examples, and a
precise cost function, each classifier's cost is computed and the minimum-cost
classifier is chosen.  However, as discussed above, such analyses assume that
the distributions are precisely known and static.
  
More general comparisons can be made with Receiver Operating Characteristic
(ROC) analysis, a classic methodology from signal detection theory that is
common in medical diagnosis and has recently begun to be used more generally
in AI classifier work
\cite{Beck-Schultz:86,Egan:75,Swets:88,FriedmanWyatt:97}.  ROC graphs depict
tradeoffs between hit rate and false alarm rate.

We use the term \textit{ROC space} to denote the coordinate system used for
visualizing classifier performance.  In ROC space, $TP$ is represented on the Y
axis and $FP$ is represented on the X axis.  Each classifier is represented by
the point in ROC space corresponding to its $(FP,TP)$ pair.  For models that
produce a continuous output, e.g., posterior probabilities, $TP$ and $FP$ vary
together as a threshold on the output is varied between its extremes (each
threshold defines a classifier); the resulting curve is called the ROC curve.
An ROC curve illustrates the error tradeoffs available with a given model.
Figure~\ref{fig:ROC-curves} shows a graph of three typical ROC curves; in fact,
these are the complete ROC curves of the classifiers shown in
figure~\ref{fig:NP}.

For orientation, several points on an ROC graph should be noted.  The lower
left point $(0,0)$ represents the strategy of never alarming, the upper right
point $(1,1)$ represents the strategy of always alarming, the point $(0,1)$
represents perfect classification, and the line $y=x$ (not shown) represents
the strategy of randomly guessing the class.  Informally, one point in ROC
space is better than another if it is to the northwest ($TP$ is higher, $FP$ is
lower, or both).  An ROC graph allows an informal visual comparison of a set of
classifiers.

ROC graphs illustrate the behavior of a classifier \emph{without
regard to class distribution or error cost}, and so they decouple
classification performance from these factors.  Unfortunately, while
an ROC graph is a valuable visualization technique, it does a poor job
of aiding the choice of classifiers.  Only when one classifier clearly
dominates another over the entire performance space can it be declared
better.

\subsection{The ROC Convex Hull method}

In this section we combine decision analysis with ROC analysis and adapt them
for comparing the performance of a set of learned classifiers.  The method is
based on three high-level principles.  First, ROC space is used to separate
classification performance from class and cost distribution information.
Second, decision-analytic information is projected onto the ROC space.  Third,
the convex hull in ROC space is used to identify the subset of classifiers
that are potentially optimal.

\begin{figure}[tb]
  \centering
  \epsfig{file=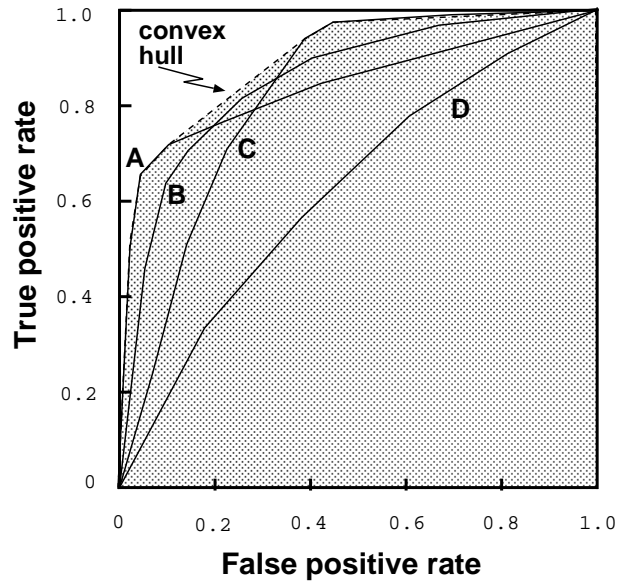}
  \caption{The ROC convex hull identifies potentially optimal classifiers.}
  \label{fig:ROC-hull}
\end{figure}

\subsubsection{Iso-performance lines}

By separating classification performance from class and cost distribution
assumptions, the decision goal can be projected onto ROC space for a neat
visualization.  Specifically, the expected cost of applying the classifier
represented by a point ($FP$,$TP$) in ROC space is:

\[
p(\POS)\cdot (1-TP)\cdot c(\NO,\POS) \; + \; p(\NEG)\cdot FP \cdot c(\YES,\NEG)
\]

Therefore, two points, ($FP_1$,$TP_1$) and ($FP_2$,$TP_2$),
have the same performance if

\[
\frac{TP_2 - TP_1}{FP_2 - FP_1} 
= 
\frac{c(\YES,\NEG)p(\NEG)}{c(\NO,\POS)p(\POS)}
\]

This equation defines the slope of an \textit{iso-performance line}.
That is, all classifiers corresponding to points on the line have the
same expected cost.  Each set of class and cost distributions defines
a family of iso-performance lines.  Lines ``more northwest'' (having a
larger $TP$-intercept) are better because they correspond to
classifiers with lower expected cost.

\subsubsection{The ROC convex hull}

Because in most real-world cases the target distributions are not known
precisely, it is valuable to be able to identify those classifiers that
potentially are optimal.  Each possible set of distributions defines a family
of iso-performance lines, and for a given family, the optimal methods are
those that lie on the ``most-northwest'' iso-performance line.  Thus, a
classifier is optimal for some conditions if and only if it lies on the
northwest boundary (\ie above the line $y=x$) of the convex hull
\cite{quickhull:96} of the set of points in ROC space.\footnote{The convex
  hull of a set of points is the smallest convex set that contains the
  points.}  We discuss this in detail in Section~\ref{sect:rocch-hybrid}.

\begin{figure}[tb]
  \centering
  \epsfig{file=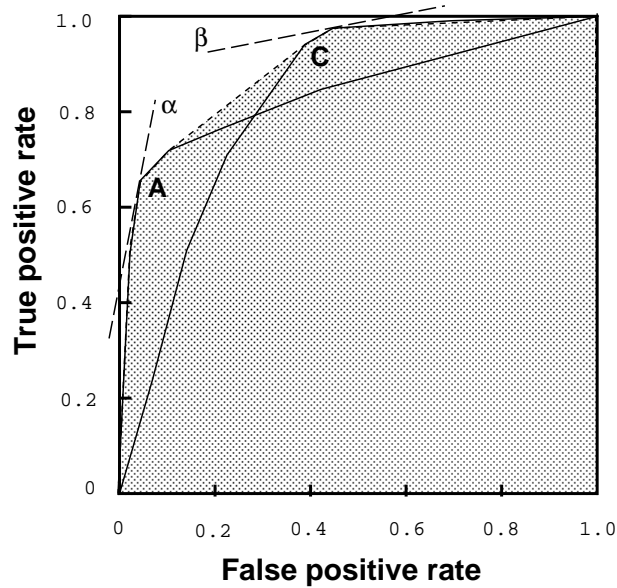}
  \caption{Lines $\alpha$ and $\beta$ show the optimal classifier under
    different sets of conditions.}
  \label{fig:ROC-hull2}
\end{figure}

We call the convex hull of the set of points in ROC space the \textit{ROC
convex hull} (\rocch) of the corresponding set of classifiers.
Figure~\ref{fig:ROC-hull} shows four ROC curves with the ROC convex hull drawn
as the border between the shaded and unshaded areas.  $\mathsf{D}$ is clearly
not optimal.  Perhaps surprisingly, $\mathsf{B}$ can never be optimal either
because none of the points of its ROC curve lies on the convex hull.  We can
also remove from consideration any points of $\mathsf{A}$ and $\mathsf{C}$
that do not lie on the hull.

Consider these classifiers under two distribution scenarios.  In each, negative
examples outnumber positives by 5:1.  In scenario $\mathcal{A}$, false
positive and false negative errors have equal cost.  In scenario $\mathcal{B}$,
a false negative is 25 times as expensive as a false positive (\eg missing a
case of fraud is much worse than a false alarm).  Each scenario defines a
family of iso-performance lines.  The lines corresponding to scenario
$\mathcal{A}$ have slope 5; those for $\mathcal{B}$ have slope $\frac{1}{5}$.
Figure~\ref{fig:ROC-hull2} shows the convex hull and two iso-performance
lines, $\alpha$ and $\beta$.  Line $\alpha$ is the ``best'' line
with slope $5$ that intersects the convex hull; line $\beta$ is the best line
with slope $\frac{1}{5}$ that intersects the convex hull.  Each line
identifies the optimal classifier under the given distribution.

\begin{figure}[tb]
  \begin{center}
    \epsfig{file=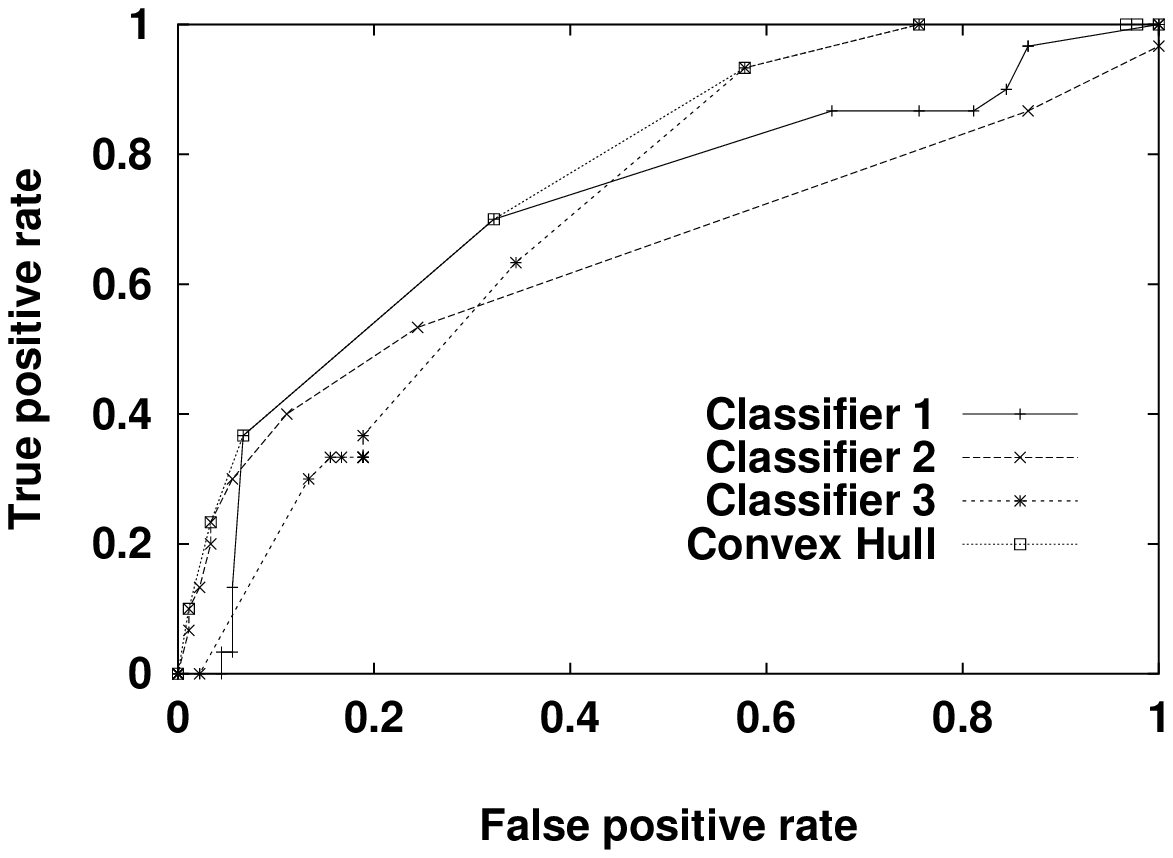,height=3in,width=3.2in}
    \caption{ROC curves with convex hull}
    \label{fig:ROCCH}
  \end{center}
\end{figure}

Figure~\ref{fig:ROCCH} shows the three ROC curves from our initial
example, with the convex hull drawn.

\subsubsection{Generating the ROC Convex Hull}

The {\it ROC convex hull method} selects the potentially optimal classifiers
based on the ROC convex hull and iso-performance lines.

\begin{table}[tb]
  \caption{Algorithm for generating an ROC curve from a set of 
    ranked examples.}
  \begin{center}
    \rule{\textwidth}{.01in}
    \begin{tabbing}
      \textbf{\rmfamily Given:}~~ \=E: \= List of \=tuples
      $\langle I, p \rangle$ where:\\
      \>\>\>$I$: labeled example\\
      \>\>\>$p$: numeric ranking assigned to $I$ by the classifier \\
      \>$P, N$: count of positive and negative examples in E, respectively.\\
      \textbf{\rmfamily Output:}  R: List of points on the ROC curve.\\
      \vspace*{1ex}\\
      xxxxx\=xxxxx\=xxxxx\=xxxxx\=xxxxx\=xxxxx\=xxxxx\=xxxxx\=xxxxx\=\kill
      $Tcount = 0$; \>\>\>\>\>\>{\it /* current TP tally */ }\\
      $Fcount = 0$; \>\>\>\>\>\>{\it /* current FP tally */ }\\
      $plast = -\infty$; \>\>\>\>\>\>{\it /* last score seen */ }\\
      $R = \langle \rangle$; \>\>\>\>\>\>{\it /* list of ROC points */ }\\
      sort $E$ in decreasing order by $p$ values;\\
      \WHILE (E $\neq \emptyset$) \DO \\
      \>remove tuple $\langle I, p \rangle$ from head of E;\\
      \>\IF ($p \neq plast$) \THEN\\
      \>\>add point ($\frac{Fcount}{N}$, $\frac{Tcount}{P}$) to end of R;\\
      \>\>$plast = p$;\\
      \>\ENDIF\\
      \>\IF ($I$ is a positive example) \THEN\\
      \>\>$Tcount = Tcount + 1$;\\
      \>\ELSE \>\>\>\>\>{\it /* I is a negative example */}\\
      \>\>$Fcount = Fcount + 1$;\\
      \>\ENDIF\\
      \ENDWHILE\\
      add point ($\frac{Fcount}{N}$, $\frac{Tcount}{P}$) to end of R;\\
    \end{tabbing}
    \rule{\textwidth}{.01in}
  \end{center}
  \label{tab:ROC-alg}
\end{table}

\begin{enumerate}
  
\item For each classifier, plot $TP$ and $FP$ in ROC space.  For
continuous-output classifiers, vary a threshold over the output range
and plot the ROC curve.  Table~\ref{tab:ROC-alg} shows an algorithm
for producing such an ROC curve in a single pass.\footnote{There is a
subtle complication to producing ROC curves from ranked test-set data,
which is reflected in the algorithm shown in Table~\ref{tab:ROC-alg}.
Specifically, consecutive examples with the same score can give overly
optimistic or overly pessimistic ROC curves, depending on the ordering
of positive and negative examples.  The ROC curve generating algorithm
shown here waits until all examples with the same score have been
tallied before computing the next point of the ROC curve.  The result
is a segment that bisects the area that would have resulted from the
most optimistic and most pessimistic orderings.}
  
\item Find the convex hull of the set of points representing the predictive
  behavior of all classifiers of interest, for example by using the QuickHull
  algorithm \cite{quickhull:96}.
  
\item For each set of class and cost distributions of interest, find the slope
  (or range of slopes) of the corresponding iso-performance lines.
  
\item For each set of class and cost distributions, the optimal classifier will
  be the point on the convex hull that intersects the iso-performance line with
  largest $TP$-intercept.  Ranges of slopes specify hull segments.

\end{enumerate}

Figures~\ref{fig:ROC-hull} and \ref{fig:ROC-hull2} demonstrate how the
subset of classifiers that are potentially optimal can be identified
and how classifiers can be compared under different cost and class
distributions.  

\subsubsection{Comparing a variety of classifiers}

\begin{figure}[tb]
  \centering \epsfig{file=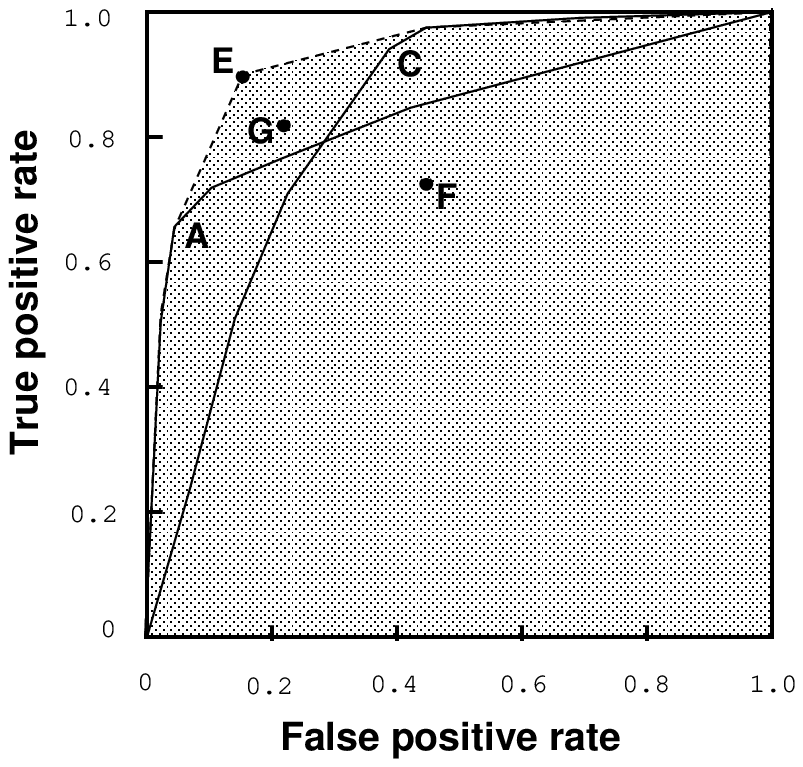,height=3in}
  \caption{Classifier $\mathsf{E}$ may be optimal for some conditions because
    it extends the ROC convex hull.  $\mathsf{F}$ and $\mathsf{G}$ cannot be
    optimal they are not on the hull, nor do they extend it.}
  \label{fig:Adding-EFG}
\end{figure}

The ROC convex hull method accommodates both binary and continuous
classifiers.  Binary classifiers are represented by individual points in ROC
space.  Continuous classifiers produce numeric outputs to which thresholds can
be applied, yielding a series of $(FP, TP)$ pairs forming an ROC curve.  Each
point may or may not contribute to the ROC convex hull.
Figure~\ref{fig:Adding-EFG} depicts the binary classifiers $\mathsf{E}$,
$\mathsf{F}$ and $\mathsf{G}$ added to the previous hull.  $\mathsf{E}$ may be
optimal under some circumstances because it extends the convex hull.
Classifiers $\mathsf{F}$ and $\mathsf{G}$ never will be optimal because they
do not extend the hull.

New classifiers can be added incrementally to an \rocch\ analysis, as
demonstrated in figure~\ref{fig:Adding-EFG} by the addition of classifiers
$\mathsf{E}$,$\mathsf{F}$, and $\mathsf{G}$.  Each new classifier either
extends the existing hull or it does not.  In the former case the hull must be
updated accordingly, but in the latter case the new classifier can be ignored.
Therefore, the method does not require saving every classifier (or saving
statistics on every classifier) for re-analysis under different
conditions---only those points on the convex hull.  Recall that each point is
a classifier and might take up considerable space.  Further, the management of
knowledge about many classifiers and their statistics from many different runs
of learning programs (e.g., with different algorithms or parameter settings)
can be a substantial undertaking.  Classifiers not on the \rocch\ can never be
optimal, so they need not be saved.  Every classifier that \emph{does} lie on
the convex hull must be saved.  In Section~\ref{sect:our-study} we demonstrate
the \rocch\ in use, managing the results of many learning experiments.

\subsubsection{Changing distributions and costs}

Class and cost distributions that change over time necessitate the reevaluation
of classifier choice.  In fraud detection, costs change based on workforce and
reimbursement issues; the amount of fraud changes monthly.  With the ROC convex
hull method, comparing under a new distribution involves only calculating the
slope(s) of the corresponding iso-performance lines and intersecting them with
the hull, as shown in figure~\ref{fig:ROC-hull2}.

The ROC convex hull method scales gracefully to any degree of
precision in specifying the cost and class distributions.  If nothing
is known about a distribution, the ROC convex hull shows all
classifiers that may be optimal under any conditions.
Figure~\ref{fig:ROC-hull} showed that, given classifiers $\mathsf{A}$,
$\mathsf{B}$, $\mathsf{C}$ and $\mathsf{D}$, only $\mathsf{A}$ and
$\mathsf{C}$ can ever be optimal.  With complete information, the
method identifies the optimal classifier(s).  In
figure~\ref{fig:ROC-hull2} we saw that classifier $\mathsf{A}$ (with a
particular threshold value) is optimal under scenario $\mathcal{A}$
and classifier $\mathsf{C}$ is optimal under scenario $\mathcal{B}$.
Next we will see that with less precise information, the ROC convex
hull can show the subset of possibly optimal classifiers.

\subsubsection{Sensitivity analysis}

\begin{figure}[tb]
  \begin{center}
    \epsfig{file=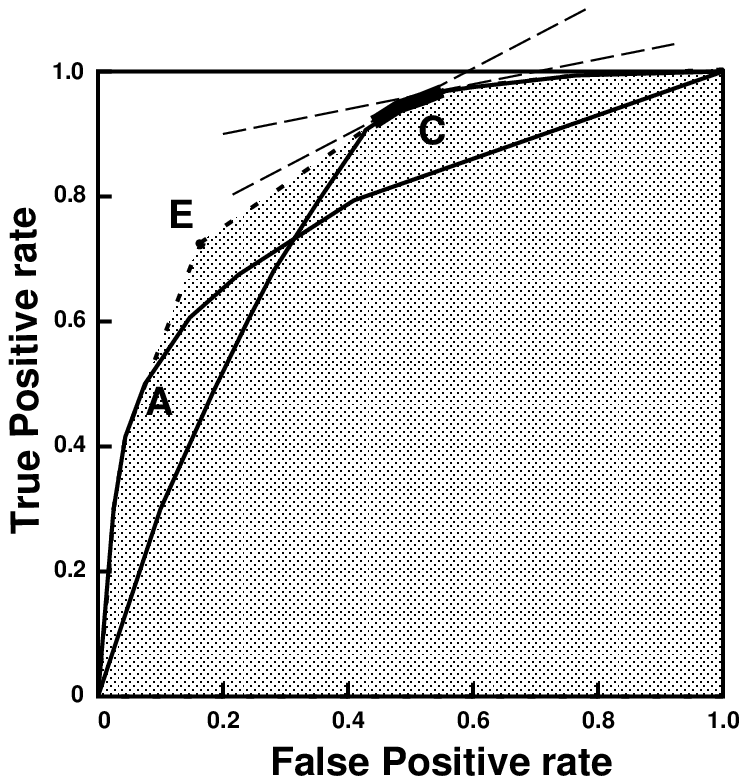,height=2.7in,width=2.6in} \\
    a.~~Low sensitivity\\
    \vspace*{.2in}
    \epsfig{file=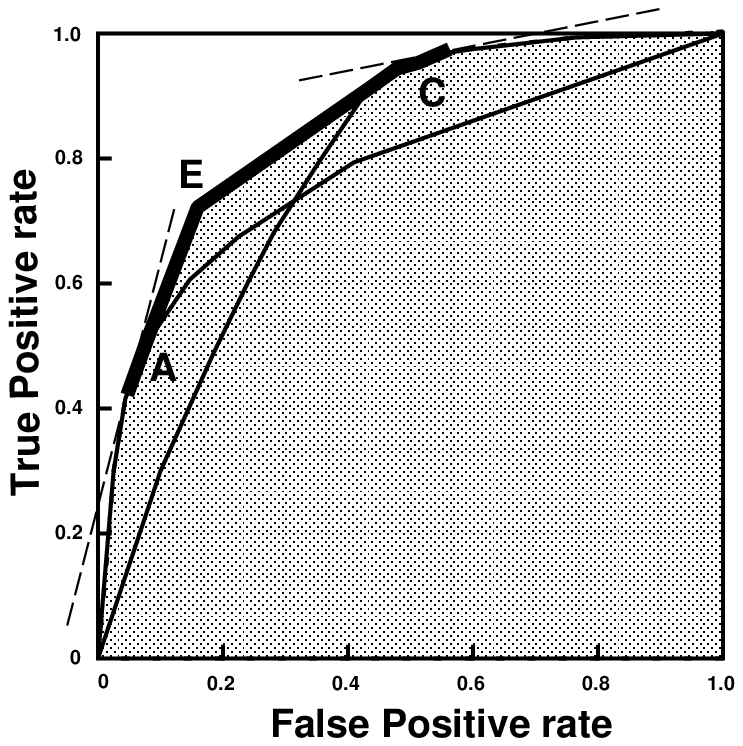,height=2.5in,width=2.5in}\\
    b.~~High sensitivity\\
  \end{center}
  \caption{Sensitivity analysis using the ROC convex hull:  (a) low
    sensitivity (only C can be optimal), (b) high sensitivity (A, E, or C can
    be optimal)}
  \label{fig:sensitive}
\end{figure}

Imprecise distribution information defines a \emph{range} of slopes for
iso-performance lines.  This range of slopes intersects a segment of the ROC
convex hull, which facilitates sensitivity analysis.  For example, if the
segment defined by a range of slopes corresponds to a single point in ROC
space or a small threshold range for a single classifier, then there is no
sensitivity to the distribution assumptions in question.  Consider a scenario
similar to $\mathcal{A}$ and $\mathcal{B}$ in that negative examples are 5
times as prevalent as positive ones.  In this scenario, consider the cost of
dealing with a false alarm to be between \$10 and \$20, and the cost of
missing a positive example to be between \$200 and \$250.  These conditions
define a range of slopes for iso-performance lines: $\frac{1}{5}\le m \le
\frac{1}{2}$.  Figure~\ref{fig:sensitive}a depicts this range of slopes and
the corresponding segment of the ROC convex hull.  The figure shows that the
choice of classifier is insensitive to changes within this range (and only
fine tuning of the classifier's threshold will be necessary).
Figure~\ref{fig:sensitive}b depicts a scenario with a wider range of slopes:
$\frac{1}{2} \le m \le 3$.  The figure shows that under this scenario the
choice of classifier is very sensitive to the distribution.  Classifiers
$\mathsf{A}$, $\mathsf{C}$ and $\mathsf{E}$ each are optimal for some
subrange.

\section{Building robust classifiers}
\label{sect:rocch-hybrid}

Up to this point, we have concentrated on the use of the \rocch\ for
visualizing and evaluating sets of classifiers.  The \rocch\ helps to
delay classifier selection as long as possible, yet provides a rich
performance comparison.  However, once system building incorporates a
particular classifier, the problem of brittleness resurfaces.  This is
important because the delay between system building and deployment may
be large, and because many systems must survive for years.  In fact,
in many domains a precise, static specification of future costs and
class distributions is not just unlikely, it is impossible
\cite{ProvostFawcettKohavi:98}.

We address this brittleness by using the \rocch\ to produce
\textbf{robust classifiers}, defined as satisfying the following.
\emph{Under any target cost and class distributions, a robust
classifier will perform at least as well as the best classifier for
those conditions.}  Our statements about optimality are practical: the
``best'' classifier may not be the Bayes-optimal classifier, but it is
at least as good as any known classifier.
Srinivasan \citeyear{Srinivasan:99} calls this ``FAPP-optimal''
(optimal for all practical purposes).  Stating that a classifier is
robust is stronger than stating that it is optimal for a specific set
of conditions.  A robust classifier is optimal under all possible
conditions.

In principle, classification brittleness could be overcome by saving
all possible classifiers (neural nets, decision trees, expert systems,
probabilistic models, etc.)  and then performing an automated run-time
comparison under the desired target conditions.  However, such a
system is not feasible because of time and space limitations---there
are myriad possible classification models, arising from the many
different learning methods under their many different parameter
settings.  Storing all the classifiers is not feasible, and tuning
the system by comparing classifiers on the fly under different
conditions is not feasible.  Fortunately, doing so is not necessary.
Moreover, we will show that it is sometimes possible to do \textit{better} than
any of these classifiers.

\subsection{ROCCH-hybrid classifiers}

We now show that robust hybrid classifiers can be built using the \rocch.

\begin{definition}
  Let $\mathbf{I}$ be the space of possible instances and let $\mathbf{C}$ be
  the space of sets of classification models.  Let a
  \mathversion{bold}$\mu$\mathversion{normal}\textbf{-hybrid classifier}
  comprise a set of classification models $\mathcal{C} \in \mathbf{C}$ and a
  function
  \[
  \mu: \mathbf{I} \times \Re \times \mathbf{C} \rightarrow \{\YES,\NO\}.
  \]
  A $\mu$-hybrid classifier takes as input an instance $I \in \mathbf{I}$ for
  classification and a number $x \in \Re$.  As output, it produces the
  classification produced by $\mu(I,x,\mathcal{C})$.
\end{definition}

Things will get more involved later, but for the time being consider that each
set of cost and class distributions defines a value for $x$, which is used to
select the (predetermined) best classifier for those conditions.  To build a
$\mu$-hybrid classifier, we must define $\mu$ and the set $\mathcal{C}$.  We
would like $\mathcal{C}$ to include only those models that perform optimally
under some conditions (class and cost distributions), since these will be
stored by the system, and we would like $\mu$ to be general enough to apply to
a variety of problem formulations.

The models comprising the {\sc rocch} can be combined to form a
$\mu$-hybrid classifier that is an elegant, robust classifier.

\begin{definition}
  The \textbf{{\sc \textbf{rocch}}-hybrid} is a $\mu$-hybrid classifier where
  $\mathcal{C}$ is the set of classifiers that form the {\sc rocch} and $\mu$
  makes classifications using the classifier on the {\sc rocch} with $FP=x$.
\end{definition}
Note that for the moment the {\sc rocch}-hybrid is defined only for $FP$
values corresponding to {\sc rocch} vertices.

\subsection{Robust classification}

Our definition of robust classifiers was intentionally vague about
what it means for one classifier to be better than another, because
different situations call for different comparison frameworks.  We now
continue with minimizing expected cost, because the process of proving
that the {\sc rocch}-hybrid minimizes expected cost for any cost and
class distributions provides a deep understanding of why and how the
{\sc rocch}-hybrid works.
Later we generalize to a wide variety of
comparison frameworks.

The \rocch-hybrid can be seen as an application of multi-criteria
optimization to classifier design and construction.  The classifiers on the
\rocch\ are Edgeworth-Pareto optimal\footnote{Edgeworth-Pareto optimality is
  the century-old notion that in a multidimensional space of criteria, optimal
  performance is the frontier of achievable performance in this space.  In
  cases where performance is a linear combination of the criteria, the
  optimality frontier is the convex hull.} \cite{Stadler-book} with respect to
TP, FP, and the objective functions we discuss.  Multi-criteria optimization
was used previously in machine learning by Tcheng, Lambert, Lu and Rendell
\shortcite{TchengEtAl:89} for the selection of inductive bias.
Alternatively, the \rocch\ can be seen as an application of the theory of
games and statistical decisions, for which convex sets (and the convex hull)
represent optimal strategies \cite{BlackwellGirshick:54}.

\subsubsection{Minimizing expected cost}

From above, the expected cost of applying a classifier is:

\begin{equation}
  \label{eq:expected_cost}
  ec(FP,TP) \; = \; p(\POS)  \cdot  (1-TP)\cdot c(\NO,\POS) \;  + 
  \; p(\NEG)  \cdot  FP \cdot c(\YES,\NEG)
\end{equation}

For a particular set of cost and class distributions, the
slope of the corresponding iso-performance lines is: 

\begin{equation}
  \label{eq:slope}
  m_{ec} = \frac{c(\YES,\NEG)p(\NEG)}{c(\NO,\POS)p(\POS)}
\end{equation}

Every set of conditions will define an $m_{ec} \ge 0$.  We now can
show that the {\sc rocch}-hybrid is robust for problems where the
``best'' classifier is the classifier with the minimum expected cost.

The slope of the {\sc rocch} is an important tool in our argument.  The {\sc
  rocch} is a piecewise-linear, concave-down ``curve.''  Therefore, as $x$
increases, the slope of the {\sc rocch} is monotonically non-increasing with
$k-1$ discrete values, where $k$ is the number of {\sc rocch} component
classifiers, including the degenerate classifiers that define the {\sc rocch}
endpoints.  Where there will be no confusion, we use phrases such as ``points
in ROC space'' as a shorthand for the more cumbersome ``classifiers
corresponding to points in ROC space.'' For this subsection, unless otherwise
noted, ``points on the
{\sc rocch}'' refer to vertices of the {\sc rocch}.

\begin{definition}
  \label{def:slope-of-rocch}
  For any real number $m \ge 0$, the \textbf{point where the slope of the
    \textsc{rocch}\ is $\mathbf{m}$} is one of the (arbitrarily chosen)
  endpoints of the segment of the {\sc rocch} with slope $m$, if such a
  segment exists.  Otherwise, it is the vertex for which the left adjacent
  segment has slope greater than $m$ and the right adjacent segment has slope
  less than $m$.
\end{definition}

For completeness, the leftmost endpoint of the {\sc rocch} is considered to be
attached to a segment with infinite slope and the rightmost endpoint of the
{\sc rocch} is considered to be attached to a segment with zero slope.  Note
that every $m \ge 0$ defines at least one point on the {\sc rocch}.

\begin{lemma}
  For any set of cost and class distributions, there is a point on the \rocch\ 
  with minimum expected cost.\\
  \textbf{Proof:} (by contradiction) Assume that for some conditions
  there exists a point \textbf{C} with smaller expected cost than any
  point on the {\sc rocch}.  By equations~\ref{eq:expected_cost} and
  \ref{eq:slope}, a point ($FP_2$,$TP_2$) has the same expected cost
  as a point ($FP_1$,$TP_1$) if \[ \frac{TP_2 - TP_1}{FP_2 - FP_1} =
  m_{ec} \] Therefore, for conditions corresponding to $m_{ec}$, all
  points with equal expected cost form an iso-performance line in ROC
  space with slope $m_{ec}$.  Also by~\ref{eq:expected_cost}
  and~\ref{eq:slope}, points on lines with larger y-intercept have
  lower expected cost.  Now, point \textbf{C} is not on the {\sc
  rocch}, so it is either above the curve or below the curve.  If it
  is above the curve, then the {\sc rocch} is not a convex set
  enclosing all points, which is a contradiction.  If it is below the
  curve, then the iso-performance line through \textbf{C} also
  contains a point \textbf{P} that is on the {\sc rocch} (not
  necessarily a vertex).  If this iso-performance line intersects no
  {\sc rocch} vertex, then consider the vertices at the endpoints of
  the {\sc rocch} segment containing \textbf{P}; one of these vertices
  must intersect a better iso-performance line than does \textbf{C}.
  In either case, since all points on an iso-performance line have the
  same expected cost, point \textbf{C} does not have smaller expected
  cost than all points on the {\sc rocch}, which is also a
  contradiction.  \EndProof
\end{lemma}

Although it is not necessary for our purposes here, it can be shown
that \textit{all} of the minimum expected-cost classifiers are
\textit{on} the {\sc rocch}.

\begin{definition}
  \label{def:m_iso_perf_line}
  An iso-performance line with slope $m$ is an \textbf{m-iso-performance
    line}.
\end{definition}

\begin{lemma}
  For any cost and class distributions that translate to $m_{ec}$, a point on
  the {\sc rocch} has minimum expected cost only if the slope
  of the {\sc rocch} at that point is $m_{ec}$.\\
  \textbf{Proof:} (by contradiction) Suppose that there is a point \textbf{D}
  on the {\sc rocch} where the slope is \emph{not} $m_{ec}$, but the point
  does have minimum expected cost.  By Definition~\ref{def:slope-of-rocch},
  either (a) the segment to the left of \textbf{D} has slope less than
  $m_{ec}$, or (b) the segment to the right of \textbf{D} has slope greater
  than $m_{ec}$.  For case (a), consider point \textbf{N}, the vertex of the
  {\sc rocch} that neighbors \textbf{D} to the left, and consider the
  (parallel) $m_{ec}$-iso-performance lines $l_D$ and $l_N$ through \textbf{D}
  and \textbf{N}.  Because \textbf{N} is to the left of \textbf{D} and the
  line connecting them has slope less than $m_{ec}$, the y-intercept of $l_N$
  will be greater than the y-intercept of $l_D$.  This means that \textbf{N}
  will have lower expected cost than \textbf{D}, which is a contradiction.
  The argument for (b) is analogous (symmetric). \EndProof
\end{lemma}

\begin{lemma}
  If the slope of the {\sc rocch} at a point is $m_{ec}$, then the point has
  minimum expected cost.\\
  \textbf{Proof:} If this point is the only point where the slope of the {\sc
    rocch} is $m_{ec}$, then the proof follows directly from Lemma 1 and
  Lemma 2.  If there are multiple such points, then by definition they are
  connected by an $m_{ec}$-iso-performance line, so they have the same
  expected cost, and once again the proof follows directly from Lemma 1 and
  Lemma 2. \EndProof
\end{lemma}

It is straightforward now to show that the {\sc rocch}-hybrid is robust for the
problem of minimizing expected cost.

\begin{theorem}
  The {\sc rocch}-hybrid minimizes expected cost for any cost distribution
  and any class distribution.\\
  \textbf{Proof:} Because the {\sc rocch}-hybrid is composed of the
  classifiers corresponding to the points on the {\sc rocch}, this follows
  directly from Lemmas 1, 2, and 3. \EndProof
\end{theorem}

Now we have shown that the {\sc rocch}-hybrid is robust when the goal
is to provide the minimum expected-cost classification.  This result
is important even for accuracy maximization, because the preferred
classifier may be different for different target class distributions.
This rarely is taken into account in experimental comparisons of
classifiers.

\begin{corollary}
  The {\sc rocch}-hybrid minimizes error rate (maximizes accuracy) for any
  target class distribution.\\
  \textbf{Proof:} Error rate minimization is cost minimization with uniform
  error costs. \EndProof
\end{corollary}

\subsection{Robust classification for other common metrics}

Showing that the \rocch-hybrid is robust not only helps us with understanding
the \rocch\ method generally, it also shows us how the \rocch-hybrid will pick
the best classifier in order to produce the best classifications, which we
will return to later.  If we ignore the need to specify how to pick the best
component classifier, we can show that the \rocch\ applies more generally.

\begin{theorem}
  \label{theorem:general-rocch}
  For any classifier evaluation metric $f(FP,TP)$, if\\
  $\Partial{f}{TP}~\ge~0$ and $\Partial{f}{FP} \le 0$ then there exists a
  point on the \rocch\ with an $f$-value at least
  as high as that of any known classifier.\\
  \textbf{Proof:} (by contradiction) Assume that there exists a classifier
  $\mathcal{C}_o$, not on the \rocch, with an $f$-value higher than that of
  any point on the \rocch.  $\mathcal{C}_o$ is either (i) above or (ii) below
  the \rocch.  In case (i), the \rocch\ is not a convex set enclosing all the
  points, which is a contradiction.  In case (ii), let $\mathcal{C}_o$ be
  represented in ROC-space by $(FP_o,TP_o)$.  Because $\mathcal{C}_o$ is below
  the \rocch\ there exist points, call one $(FP_p,TP_p)$, on the \rocch\ with
  $TP_p > TP_o$ and $FP_p < FP_o$.  However, by the restriction on the partial
  derivatives, for any such point $f(FP_p,TP_p) \ge f(FP_o,TP_o)$, which again
  is a contradiction.  \EndProof
\end{theorem}

There are two complications to the more general use of the \rocch,
both of which are illustrated by the decision criterion from our very
first example.  Recall that the Neyman-Pearson criterion specifies a
maximum acceptable $FP$ rate.  Standard ROC analysis uses ROC curves
to select a single, parameterized classification model; the parameter
allows the user to select the ``operating point'' for a
decision-making task, usually a threshold on a probabilistic output
that will allow for optimal decision making.  Under the Neyman-Pearson
criterion, selecting the single best model from a set is easy: plot
the ROC curves, draw a vertical line at the desired maximum $FP$, and
pick the model whose curve has the largest $TP$ at the intersection
with this line.

\begin{figure}[tb]
  \begin{center}
    \epsfig{file=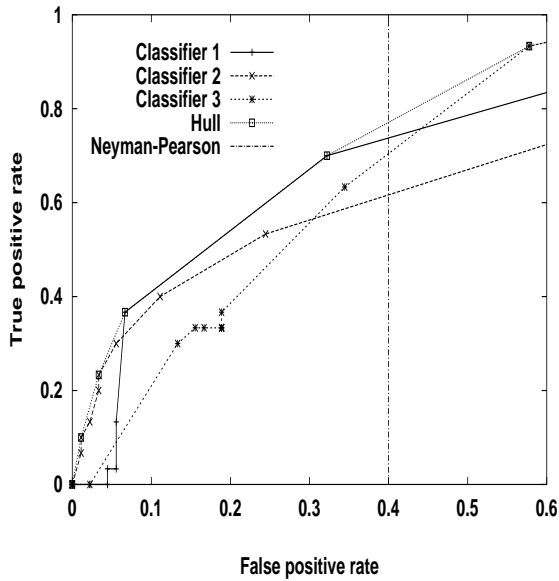,height=3.1in,width=3in}
    \caption{The ROC Convex Hull used to select a classifier under the
      Neyman-Pearson criterion}
    \label{fig:ROC-NP}
  \end{center}
\end{figure}

With the \rocch-hybrid, making the best classifications under
the Neyman-Pearson criterion is not so straightforward.
For minimizing expected cost it was sufficient for the {\sc rocch}-hybrid to
choose a \textit{vertex} from the {\sc rocch} for any $m_{ec}$ value.  For
problem formulations such as the Neyman-Pearson criterion, the performance
statistics at a non-vertex point on the {\sc rocch} may be preferable (see
figure~\ref{fig:ROC-NP}).  Fortunately, with a slight extension, the {\sc
  rocch}-hybrid can yield a classifier with these performance statistics.

\begin{theorem}
  \label{theorem:rocch-achieves-any-tradeoff} An {\sc rocch}-hybrid
  can achieve the $TP$:$FP$ tradeoff represented by any point on the
  {\sc rocch}, not just the vertices.\\ \textbf{Proof:} (by
  construction) Extend $\mu(I,x,\mathcal{C})$ to non-vertex points as
  follows.  Pick the point $P$ on the {\sc rocch} with $FP=x$ (there
  is exactly one).  Let $TP_x$ be the $TP$ value of this point.  If
  ($x$, $TP_x$) is an {\sc rocch} vertex, use the corresponding
  classifier.  If it is not a vertex, call the left endpoint of the
  hull segment on which $P$ lies $C_l$, and the right endpoint $C_r$.
  Let $d$ be the distance between $C_l$ and $C_r$, and let $p$ be the
  distance between $C_l$ and $P$.  Make classifications as follows.
  For each input instance flip a weighted coin and choose the answer
  given by classifier $C_r$ with probability $\frac{p}{d}$ and that
  given by classifier $C_l$ with probability $1-\frac{p}{d}$.  It is
  straightforward to show that $FP$ and $TP$ for this classifier will
  be $x$ and $TP_x$. \EndProof
\end{theorem}

The second complication is that, as illustrated by the Neyman-Pearson
criterion, many practical classifier comparison frameworks include
\textit{constrained} optimization problems (below we will discuss other
frameworks).  Arbitrarily constrained optimizations are problematic for the
\rocch-hybrid.  Given total freedom, it is possible to devise constraints on
classifier performance such that, even with the restriction on the partial
derivatives, an interior point scores higher than any \textit{acceptable}
point on the hull.  For example, two linear constraints can enclose a subset
of the interior and exclude \textit{the entire} \rocch---there will be no
acceptable points on the \rocch.  However, many realistic constraints do not
thwart the optimality of the \rocch-hybrid.

\begin{theorem}
  \label{theorem:general-rocch-hybrid}
  For any classifier evaluation metric $f(FP,TP)$, if \\
  $\Partial{f}{TP}\ge~0$ and $\Partial{f}{FP}\le~0$ and no constraint on
  classifier performance eliminates any point on the \rocch\ without also
  eliminating all higher-scoring interior points, then the \rocch-hybrid can
  perform at least as well as any known classifier.
  \\
  \textbf{Proof:} Follows directly from Theorem~\ref{theorem:general-rocch}
  and Theorem~\ref{theorem:rocch-achieves-any-tradeoff}.  \EndProof
\end{theorem}

Linear constraints on classifiers' $FP:TP$ performance are common
for real-world problems, so the following is
useful.

\begin{corollary}
  \label{corollary:linear-constraints}
  For any classifier evaluation metric $f(FP,TP)$, if\\
  $\Partial{f}{TP} \ge 0$ and $\Partial{f}{FP} \le 0$
  and there is a single constraint on classifier performance
  of the form $a \cdot TP + b \cdot FP \le c$, with $a$ and $b$
  non-negative,
  then
  the \rocch-hybrid can perform at least as well as any known
  classifier.
  \\
  \textbf{Proof:}
  The single constraint eliminates from contention all points (classifiers)
  that do not fall to the left of, or below, a line with non-positive
  slope.  By the restriction on the partial derivatives, such a constraint
  will not eliminate a point on the \rocch\  without also eliminating
  all interior points with higher $f$-values.
  Thus, the proof follows directly from Theorem~\ref{theorem:general-rocch-hybrid}.
  \EndProof
\end{corollary}

So, finally, we have the following:

\begin{corollary}
  \label{cor:rocch-maximizes-NP}
  For the Neyman-Pearson criterion, the {\sc rocch}-hybrid can perform at
  least as well as that of any known
  classifier.\\
  \textbf{Proof:} For the Neyman-Pearson criterion, the evaluation metric is
  $f(FP,TP)=TP$, that is, a higher $TP$ is better.  The constraint on
  classifier performance is $FP \le FP_{max}$. These satisfy the conditions
  for Corollary~\ref{corollary:linear-constraints}, and therefore this
  corollary follows.  \EndProof
\end{corollary}

All the foregoing effort may seem misplaced for a simple
criterion like Neyman-Pearson.  However, there are
many other realistic problem formulations.  
For example, consider
the decision-support problem of optimizing \textit{workforce utilization}, in
which a workforce is available that can process a fixed number of cases.  Too few
cases will under-utilize the workforce, but too many cases will leave some
cases unattended (expanding the workforce usually is not a short-term
solution).  If the workforce can handle $K$ cases, the system should present
the best possible set of $K$ cases.  This is similar to the Neyman-Pearson
criterion, but with an absolute cutoff ($K$) instead of a percentage cutoff
($FP$).

\begin{theorem}
  \label{the:rocch_best}
  For workforce utilization, the {\sc rocch}-hybrid will provide the best set
  of $K$ cases, for any choice of $K$.\\ 
 \textbf{Proof:} (by construction) The decision criterion is to maximize $TP$
  subject to the constraint:
  \[
  TP \cdot P + FP \cdot N \le K
  \]
  The theorem therefore follows from Corollary~\ref{corollary:linear-constraints}. \EndProof
\end{theorem}

In fact, many screening problems, such as are found in marketing and
information retrieval, use exactly this linear constraint.  It follows that
for maximizing lift \cite{BerryLinoff:97}, precision, or recall, subject to
absolute or percentage cutoffs on case presentation, the {\sc rocch}-hybrid
will provide the best set of cases.

As with minimizing expected cost, imprecision in the environment
forces us to favor a \textit{robust} solution for these other
comparison frameworks.  For many real-world problems, the precise
desired cutoff will be unknown or will change (\eg because of
fundamental uncertainty, variability in case difficulty, or competing
responsibilities).  What is worse, for a fixed (absolute) cutoff
merely changing the size of the universe of cases (e.g., the size of
a document corpus) may change the preferred classifier, because it
will change the constraint line.  The {\sc rocch}-hybrid provides a
robust solution because it gives the optimal subset of cases for any
constraint line.  For example, for document retrieval the {\sc
rocch}-hybrid will yield the best $N$ documents for any $N$, for any
prior class distribution (in the target corpus), and for any target
corpus size.

\subsection{Ranking cases}
\label{sect:ranking-cases}

An apparent solution to the problem of robust classification is to use a model
that ranks cases, and just work down the ranked list.  This approach appears
to sidestep the brittleness demonstrated with binary classifiers, since the
choice of a cutoff point can be deferred to classification time.  However,
choosing the best ranking model is still problematic.  For most practical
situations, choosing the best ranking model is equivalent to choosing which
classifier is best \emph{for the cutoff that will be used}.

\begin{figure}[tb]
  \begin{center}
    \epsfig{file=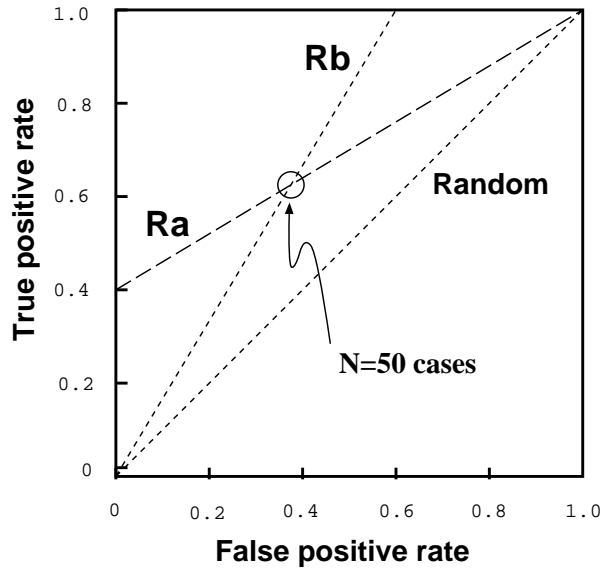,height=3in}
    \caption{The ROC curves of the two ranking classifiers, $R_a$ and $R_b$,
      described in Section~\ref{sect:ranking-cases}.}
    \label{fig:Ranking-models}
  \end{center}
\end{figure}

An example will illustrate this.  Consider two ranking functions, $R_a$ and
$R_b$, applied to a class-balanced set of 100 cases.  Assume $R_a$ is able to
recognize a common aspect unique to positive cases that occurs in 20\% of the
population, and it ranks these highest.  Assume $R_b$ is able to recognize a
common aspect unique to negative cases occurring in 20\% of the population, and it
ranks these lowest.  So $R_a$ ranks the highest 20\% correctly and performs
randomly on the remainder, while $R_b$ ranks the lowest 20\% correctly and
performs randomly on the remainder.  Which model is better?  The answer
depends entirely upon how far down the list the system will go before it
stops; that is, upon what cutoff will be used.  If fewer than 50 cases are to
be selected then $R_a$ should be used, whereas $R_b$ is better if more than 50
cases will be selected.  Figure~\ref{fig:Ranking-models} shows the ROC curves
corresponding to these two classifiers, and the point corresponding to $N=50$
where the curves cross in ROC space.

The \rocch\ method can be used to organize such ranking models, as we have
seen.  Recall that ROC curves are formed from case rankings by moving the
cutoff from one extreme to the other (Table~\ref{tab:ROC-alg} shows an
algorithm for calculating the ROC curve from such rankings).  The {\sc
  rocch}-hybrid comprises the ranking models that are best for all possible
conditions.

\subsection{Whole-curve metrics}

In situations where either the target cost distribution or class distribution
is \emph{completely} unknown, some researchers advocate choosing the
classifier that maximizes a single-number metric representing the average
performance over the entire curve.  A common whole-curve metric is ``AUC'',
the Area Under the (ROC) Curve \cite{Bradley:97}.  The AUC is equivalent to
the probability that a randomly chosen positive instance will be rated higher
than a negative instance, and thereby is also estimated by the Wilcoxon test
of ranks \cite{HanleyMcNeil:82}.  A criticism of AUC is that for specific
target conditions the classifier with the maximum AUC may be suboptimal
\cite{ProvostFawcettKohavi:98}.  Indeed, this criticism may be made of any
single-number metric.  Fortunately, not only is the \textsc{rocch}-hybrid
optimal for any specific target conditions, it has the maximum 
AUC---There is no classifier with AUC larger than that of the {\sc rocch}-hybrid.

\subsection{Using the ROCCH-hybrid}

To use the \textsc{rocch}-hybrid for classification, we need to translate
environmental conditions to $x$ values to plug into $\mu(I,x,\mathcal{C})$.
For minimizing expected cost, Equation~\ref{eq:slope} shows how to translate
conditions to $m_{ec}$.  For any $m_{ec}$, by Lemma~3 we want the $FP$ value
of the point where the slope of the {\sc rocch} is $m_{ec}$, which is
straightforward to calculate.  For the Neyman-Pearson criterion the conditions
are defined as $FP$ values.  For workforce utilization with conditions
corresponding to a cutoff $K$, the $FP$ value is found by intersecting the line
$TP \cdot P + FP \cdot N = K$ with the {\sc rocch}.

We have argued that target conditions (misclassification costs and
class distribution) are rarely known.  It may be confusing that
we now seem to require exact knowledge of these conditions.  The
\textsc{rocch}-hybrid gives us two important capabilities.  First, the
need for precise knowledge of target conditions is deferred until
run time.  Second, in the absence of precise knowledge even at
run time, the system can be optimized easily with minimal feedback.

By using the \textsc{rocch}-hybrid, information on target conditions is not
needed to train and compare classifiers.  This is important because of 
imprecision caused by temporal,
geographic, or other differences that may exist between training and use.  
For example, building
a system for a real-world problem introduces a non-trivial delay between the
time data are gathered and the time the learned models will be used.  The
problem is exacerbated in domains where error costs or class distributions
change over time; even with slow drift, a brittle model may become suboptimal
quickly.  In many such scenarios, costs and class distributions can be specified
(or respecified) at run time with reasonable precision by sampling from the
current population, and used to ensure that the {\sc rocch}-hybrid always
performs optimally.

In some cases, even at run time these quantities are not known
exactly.  A further benefit of the \textsc{rocch}-hybrid is that it
can be tuned easily to yield optimal performance with only minimal
feedback from the environment.  Conceptually, the {\sc rocch}-hybrid
has one ``knob'' that varies $x$ in $\mu(I,x,\mathcal{C})$ from one
extreme to the other.  For any knob setting, the {\sc rocch}-hybrid
will give the optimal $TP$:$FP$ tradeoff for the target conditions
corresponding to that setting.  Turning the knob to the right
increases $TP$; turning the knob to the left decreases $FP$.  Because
of the monotonicity of the \textsc{rocch}-hybrid, simple hill-climbing
can guarantee optimal performance.  For example, if the system
produces too many false alarms, turn the knob to the left; if the
system is presenting too few cases, turn the knob to the right.

\subsection{Beating the component classifiers}
\label{sect:beating-the-components}

Perhaps surprisingly, in many realistic situations an {\sc
rocch}-hybrid system can do \emph{better} than any of its component
classifiers.  Consider the Neyman-Pearson decision criterion.  The
{\sc rocch} may intersect the $FP$-line \textit{above} the highest
component ROC curve.  This occurs when the $FP$-line intersects the
{\sc rocch} between vertices; therefore, there is no component
classifier that actually produces these particular ($FP$,$TP$)
statistics, as in figure~\ref{fig:ROC-NP}.  By
Theorem~\ref{theorem:rocch-achieves-any-tradeoff}, the {\sc
rocch}-hybrid can achieve any $TP$ on the hull.  Only a small number
of $FP$ values correspond to hull vertices.
The same holds for other common problem formulations, such as workforce
utilization, lift maximization, precision maximization, and recall
maximization.

\subsection{Time and space efficiency}

We have argued that the {\sc rocch}-hybrid is robust for a wide variety of
problem formulations.  It is also efficient to build, to store, and to update.

The time efficiency of building the {\sc rocch}-hybrid depends first
on the efficiency of building the component models, which varies
widely by model type.  Some models built by machine learning methods
can be built in seconds (once data are available).  Hand-built models
can take years to build.  However, we presume that this is work that
would be done anyway.  The {\sc rocch}-hybrid can be built with
whatever methods are available, be there two or two thousand. As
described below, as new classifiers become available, the {\sc
rocch}-hybrid can be updated incrementally.  The time efficiency
depends also on the efficiency of the experimental evaluation of the
classifiers.  Once again, we presume that this is work that would be
done anyway.  Finally, the time efficiency of the {\sc rocch}-hybrid
depends on the efficiency of building the {\sc rocch}, which can be
done in $O(N \log N)$ time using the QuickHull algorithm
\cite{quickhull:96} where $N$ is the number of classifiers.

The {\sc rocch} is space efficient, too, because it comprises only
classifiers that might be optimal under some target conditions (which
follows directly from Lemmas 1--3 and Definitions 3 and 4).  The
number of classifiers that must be stored can be reduced if bounds can
be placed on the potential target conditions.  As described above,
ranges of conditions define segments of the {\sc rocch}.  Thus, the
{\sc rocch}-hybrid may need only a subset of $\mathcal{C}$.

Adding new classifiers to the {\sc rocch}-hybrid also is efficient.  Adding a
classifier to the \textsc{rocch} will either (i) extend the hull, adding to
(and possibly subtracting from) the {\sc rocch}-hybrid, or (ii) conclude that
the new classifiers are not superior to the existing classifiers in any
portion of ROC space and can be discarded.

The run-time (classification) complexity of the {\sc rocch}-hybrid is never
worse than that of the component classifiers.  In situations where run-time
complexity is crucial, the {\sc rocch} should be constructed without
prohibitively expensive classification models.  It then will find the best
subset of the computationally efficient models.

\section{Empirical demonstration of need}

Robust classification is of fundamental interest because it
weakens two very strong assumptions: the
availability of precise knowledge of costs and 
of class distributions.
However, might it not be that existing classifiers already are robust?
For example, if a given classifier is optimal under one set of
conditions, might it not be optimal under all?

It is beyond the scope of this paper to offer an in-depth experimental study
answering this question.  However, we can provide solid evidence that the
answer is ``no'' by referring to the results of two prior studies.  One is a
comprehensive ROC analysis of medical domains recently conducted by Andrew
Bradley \citeyear{Bradley:97}.\footnote{Bradley's purpose was not to answer
  this question; fortunately, his published results do anyway.}  The other is a
published ROC analysis of UCI database domains that we undertook last year
with Ron Kohavi \cite{ProvostFawcettKohavi:98}.

Note that a classifier \textit{dominates} if its ROC curve completely
defines the {\sc rocch} (which means dominating classifiers are robust
and vice versa).  Therefore, if there exist more than a trivially few
domains where no single classifier dominates, then techniques like the {\sc
rocch}-hybrid are essential if robust classifiers are desired.

\subsection{Bradley's study}

Bradley studied six
medical data sets, noting that ``unfortunately, we rarely know what the
individual misclassification costs are.''  He plotted the ROC curves of six
classifier learning algorithms (two neural nets, two decision trees and two
statistical techniques).

\begin{figure}[tb]
  \begin{center}
    \epsfig{file=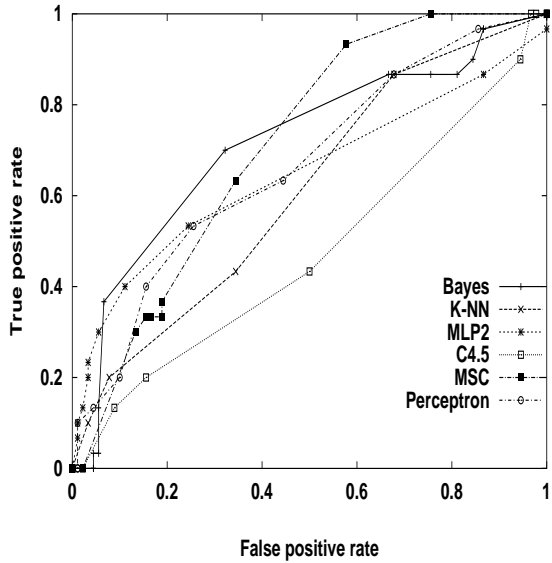,height=3in,width=3in}
    \caption{Bradley's classifier results for the heart bleeding data.}
    \label{fig:Bradley-HB}
  \end{center}
\end{figure}

\begin{figure}[tb]
  \begin{center}
    \epsfig{file=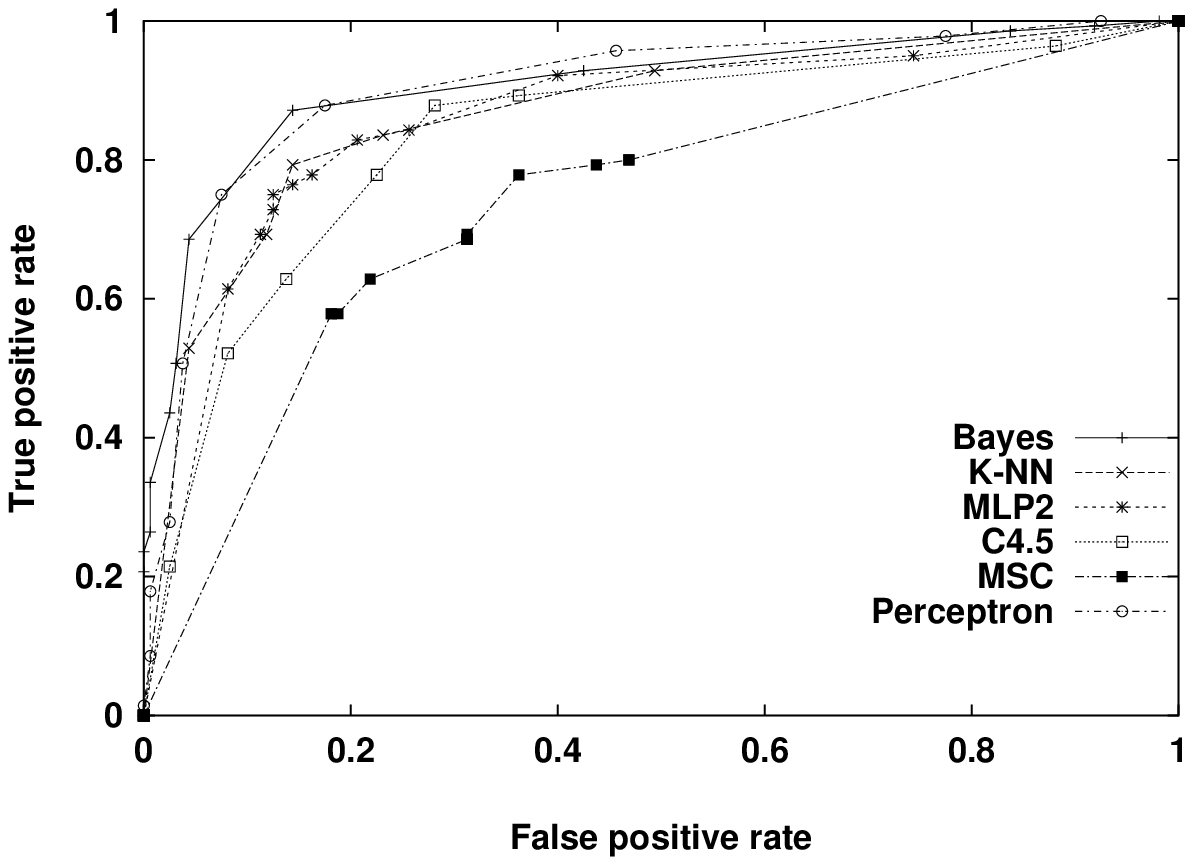,height=3in,width=3in}
    \caption{Bradley's classifier results for the Cleveland heart disease data}
    \label{fig:Bradley-Cleveland}
  \end{center}
\end{figure}

On \textit{not one} of these data sets was there a dominating
classifier.  This means that for each domain, there exist different
sets of conditions for which different classifiers are preferable.  In
fact, the running example in the present article is based on the three
best classifiers from Bradley's results on the heart bleeding data;
his results for the full set of six classifiers can be found in
figure~\ref{fig:Bradley-HB}.  Classifiers constructed for the
Cleveland heart disease data are shown in
figure~\ref{fig:Bradley-Cleveland}.

Bradley's results show clearly that for many domains the classifier that
maximizes any single metric---be it accuracy, cost, or the area under the ROC
curve---will be the best for some cost and class distributions and will not be
the best for others.  We have shown that the {\sc
  rocch}-hybrid will be the best for all.

\subsection{Our study}
\label{sect:our-study}

In the study we performed with Ron Kohavi, we chose ten datasets from the UCI
repository, each of which contains at least 250 instances, but for which the
accuracy for decision trees was less than 95\%.  For each domain, we induced
classifiers for the minority class (for Road, we chose the class Grass).  We
selected several induction algorithms from \mlc\ \cite{mlc-new-intro-j}: a
decision tree learner (MC4), Naive Bayes with discretization (NB), $k$-nearest
neighbor for several $k$ values (IB$k$), and Bagged-MC4
\cite{breiman-bagging}.  MC4 is similar to C4.5 \cite{quinlan-c45};
probabilistic predictions are made by using a Laplace correction at the
leaves.  NB discretizes the data based on entropy minimization
\cite{dougherty-kohavi-sahami-disc} and then builds the Naive-Bayes model
\cite{domingos-pazzani-simple-bayes}.  IB$k$ votes the closest $k$ neighbors;
each neighbor votes with a weight equal to one over its distance from the test
instance.

\begin{figure}[tb]
  \centerline{%
    \begin{tabular}{c@{\hspace{3pc}}c}
      \epsfig{file=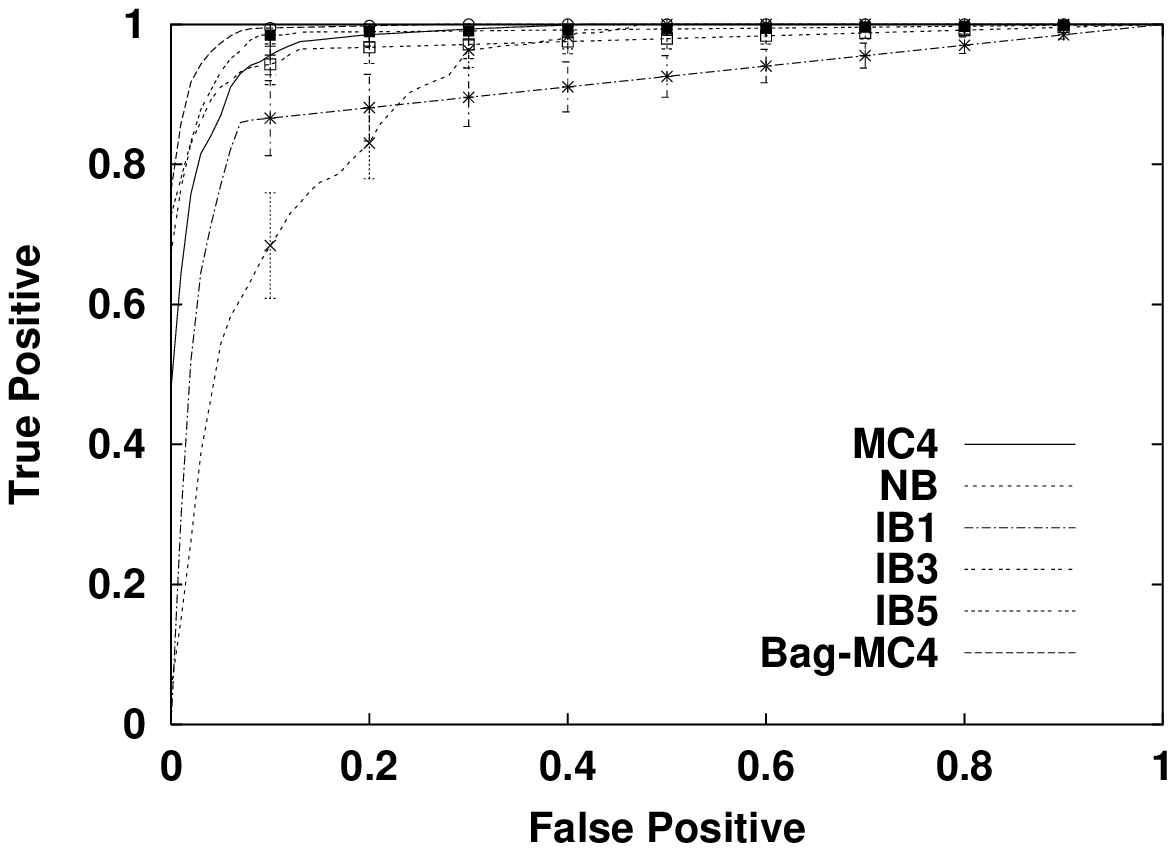,height=2.7in,width=2.7in} & 
      \epsfig{file=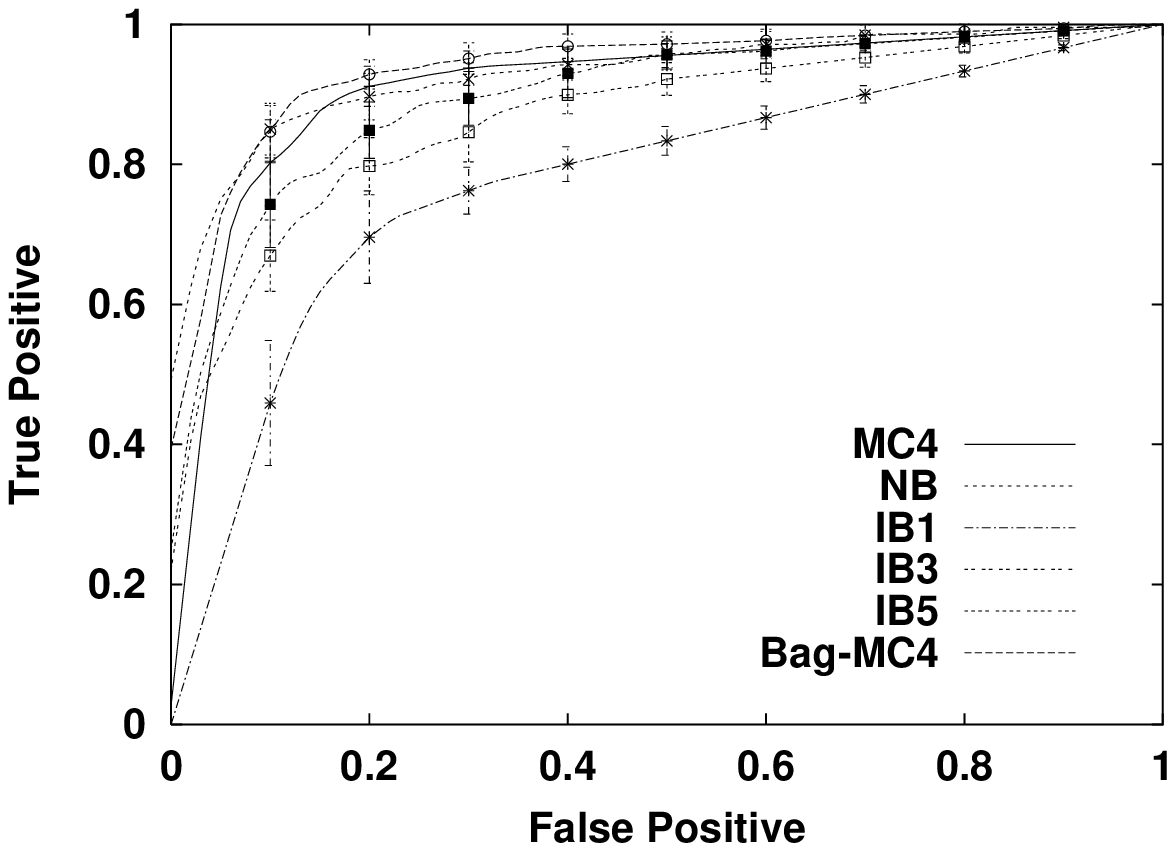,    height=2.7in,width=2.7in}\\
      a.~~Vehicle                        & 
      b.~~CRX \\
      \\
      \epsfig{file=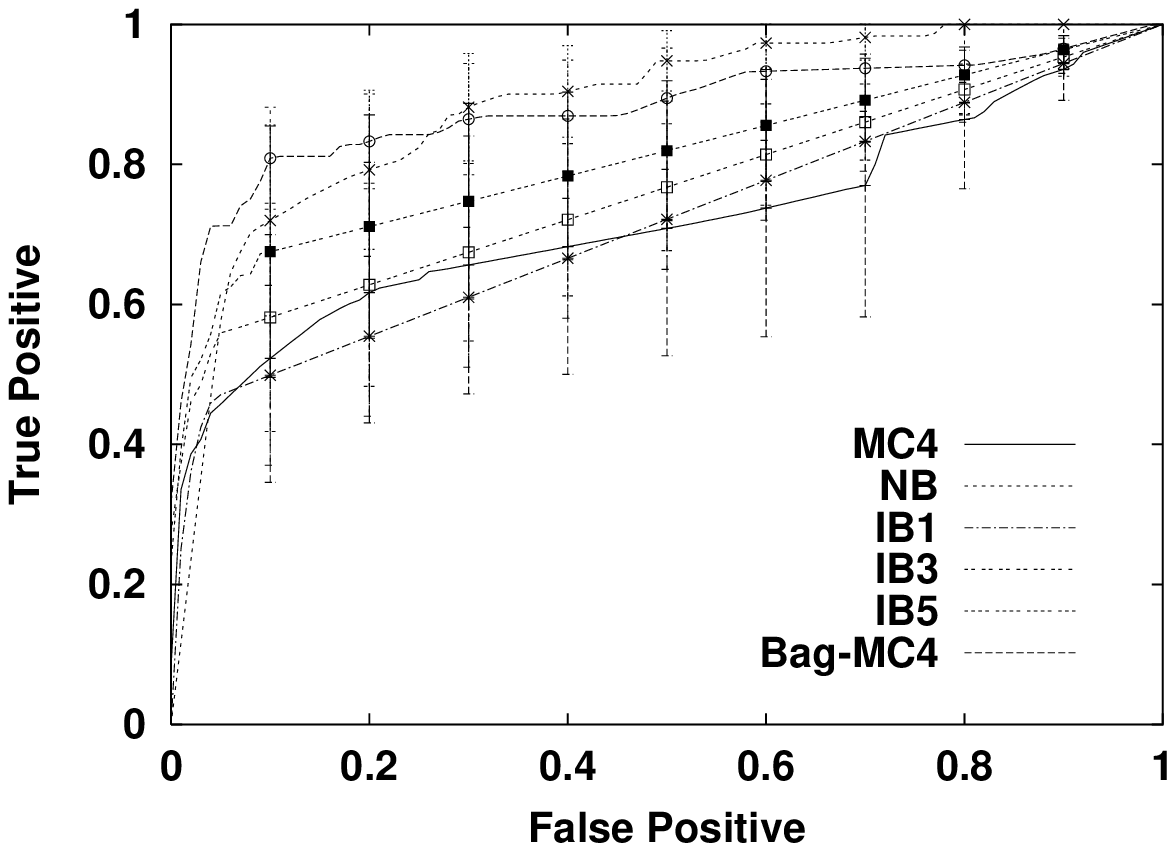,height=2.7in,width=2.7in} & 
      \epsfig{file=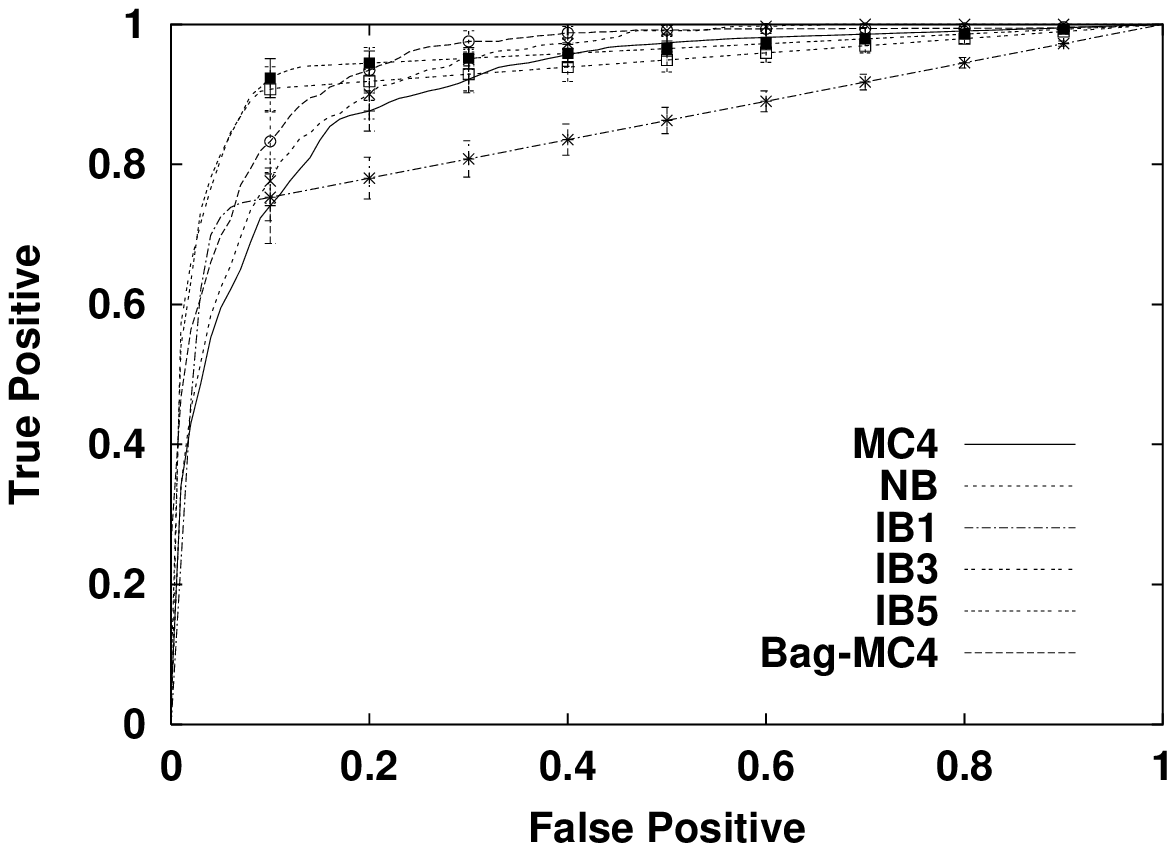, height=2.7in,width=2.7in}\\
      c.~~RoadGrass                        & 
      d.~~Satimage
    \end{tabular}
    }
  \caption{Smoothed ROC curves from UCI database domains}
  \label{fig:UCI-ROCs}
\end{figure}

Some of the ROC curves are shown in Figure~\ref{fig:UCI-ROCs}.  For \emph{only
  one} of these ten domains (Vehicle) was there an absolute dominator.  In
general, very few of the 100 runs performed (on 10 data sets, using 10
cross-validation folds each) had dominating classifiers.  Some cases are very
close, for example Adult and Waveform-21.  In other cases a curve that
dominates in one area of ROC space is dominated in another.  These results
also support the need for methods like the \rocch -hybrid, which produce
robust classifiers.

\begin{table}[tb]
  \caption{Locally dominating classifiers for four UCI domains}
  \label{tab:convex-hulls}
  \normalsize
  \begin{tabular*}{3.5in}{lll}
      \textbf{Domain} & \textbf{Slope range} & \textbf{Dominator} \\ \hline
      Vehicle         & [0, $\infty$)       & Bagged-MC4\\ \hline
      Road (Grass)    & [0, 0.38]           & NB\\
                      & [0.38, $\infty$)    & Bagged-MC4\\ \hline
      CRX             & [0, 0.03]           & Bagged-MC4\\
                      & [0.03, 0.06]        & NB\\
                      & [0.06, 2.06]        & Bagged-MC4\\
                      & [2.06, $\infty$)    & NB\\ \hline
      Satimage        & [0, 0.05]           & NB \\
                      & [0.05, 0.22]        & Bagged-MC4 \\            
                      & [0.22, 2.60]        & IB5 \\
                      & [2.60, 3.11]        & IB3 \\ 
                      & [3.11, 7.54]        & IB5 \\                       
                      & [7.54, 31.14]       & IB3 \\
                      & [31.14, $\infty$)   & Bagged-MC4 \\ \hline
  \end{tabular*}
\end{table}

As examples of what expected-cost-minimizing \textsc{rocch}-hybrids would look
like internally, Table~\ref{tab:convex-hulls} shows the component classifiers
that make up the \rocch\ for the four UCI domains of
figure~\ref{fig:UCI-ROCs}.  For example, in the Road domain (see
figure~\ref{fig:UCI-ROCs} and Table~\ref{tab:convex-hulls}), Naive Bayes would
be chosen for any target conditions corresponding to a slope less than $0.38$,
and Bagged-MC4 would be chosen for slopes greater than $0.38$.  They perform
equally well at $0.38$.

\section{Limitations and future work}

There are limitations to the {\sc rocch} method as we have presented it here.
We have defined it here only for two-class problems.  Srinivasan
\citeyear{Srinivasan:99} shows that it can be extended to multiple dimensions.
It should be noted that the dimensionality of the ``ROC-hyperspace'' grows
quadratically in the number of classes, so both efficiency and visualization
capability are called into question.

We have assumed constant error costs for a given \textit{type} of
error, e.g., all false positives cost the same.  For some problems,
different errors of the same type have different costs.  In many
cases, such a problem can be transformed for evaluation into an
equivalent problem with uniform intra-type error costs by duplicating
instances in proportion to their costs (or by simply modifying the
counting procedure accordingly).

We also have assumed for this paper that the estimates of the classifiers'
performance statistics ($FP$ and $TP$) are very good.  As mentioned above, much
work has addressed the production of good estimates for simple performance
statistics such as error rate.  Much less work has addressed the production of
good ROC curve estimates.  As with simpler statistics, care should be taken to
avoid over-fitting the training data and to ensure that differences between ROC
curves are meaningful.  One solution is to use cross-validation with averaging
of ROC curves \cite{ProvostFawcettKohavi:98}, which is the procedure used to
produce the ROC curves in Section~\ref{sect:our-study}.  To our knowledge, the
issue is open of how best to produce confidence bands appropriate to a
particular problem.  Those shown in Section~\ref{sect:our-study} are
appropriate for the Neyman-Pearson decision criterion (i.e., they show
confidence on $TP$ for various values of $FP$).

Also, we have addressed predictive performance and computational
performance.  These are not the only concerns in choosing a
classification model.  What if comprehensibility is important?  The
easy answer is that for any particular setting, the {\sc rocch}-hybrid
is as comprehensible as the underlying model it is using.  However,
this answer falls short if the {\sc rocch}-hybrid is interpolating
between two models or if one wants to understand the
``multiple-model'' system as a whole.

Although ROC analysis and the ROCCH method were specifically designed for
classification domains, we have extended them to \emph{activity monitoring}
domains \cite{FawcettProvost:99}.  Such domains involve monitoring the
behavior of a population of entities for interesting events requiring action.
These problems are substantially different from standard classification because
timeliness of classification is important and dependencies exist among
instances; both factors complicate evaluation.

This work is fundamentally different from other recent machine
learning work on combining multiple models \cite{AliPazzani:96}.  That work
combines models in order to boost performance for a fixed cost and class
distribution.  The {\sc rocch}-hybrid combines models for robustness across
different cost and class distributions.  In principle, these methods should be
independent---multiple-model classifiers are candidates for extending the {\sc
  rocch}.  However, it may be that some multiple-model classifiers achieve
increased performance for a specific set of conditions by (in effect)
interpolating along edges of the {\sc rocch}.
Cherikh \cite{Cherikh-thesis} uses
ROC analysis to study decision making where the decisions of
multiple models are present.  Unlike our work, the goal is to   
find optimal combinations of models for specific conditions.  
However, it seems that the two methods may be combined profitably:
well-chosen combinations of models
should extend the ROCCH, yielding a better robust classifier.

The \rocch\ method also complements research on cost-sensitive learning
\cite{Turney-cost-bib}.  Existing cost-sensitive learning methods are brittle
with respect to imprecise cost knowledge.  Thus, the \rocch\ is an essential
evaluation tool.  Furthermore, cost-sensitive learning may be used to find
better components for the \rocch-hybrid, by searching explicitly for
classifiers that extend the \rocch.

\section{Conclusion}

The ROC convex hull method is a robust, efficient solution to the
problem of comparing multiple classifiers in imprecise and changing
environments.  It is intuitive, can compare classifiers both in general
and under specific distribution assumptions, and provides crisp
visualizations.  It minimizes the management of classifier performance
data, by selecting exactly those classifiers that are potentially
optimal; thus, only these need to be saved in preparation for
changing conditions.  Moreover, due to its incremental nature, new
classifiers can be incorporated easily, \eg when trying a new parameter
setting.

The {\sc rocch}-hybrid performs optimally under any target conditions
for many realistic problem formulations, including the optimization of
metrics such as accuracy, expected cost, lift, precision, recall, and
workforce utilization.  It is efficient to build in terms of time and
space, and can be updated incrementally.  Furthermore, it can
sometimes classify better than any (other) known model.  Therefore, we
conclude that it is an elegant, robust classification system.

We believe that this work has important implications for both machine learning
applications and machine learning research \cite{ProvostFawcettKohavi:98}.  For
applications, it helps free system designers from the need to choose (sometimes
arbitrary) comparison metrics before precise knowledge of key evaluation
parameters is available.  Indeed, such knowledge may never be available, yet
robust systems still can be built.

For machine learning research, it frees researchers from the need to
have precise class and cost distribution information in order to study
important related phenomena.  In particular, work on cost-sensitive
learning has been impeded by the difficulty of specifying costs, and
by the tenuous nature of conclusions based on a single cost metric.
Researchers need not be held back by either.  Cost-sensitive learning
can be studied generally without specifying costs precisely.  The same
goes for research on learning with highly skewed distributions.  Which
methods are effective for which levels of distribution skew?  The
\rocch\ will provide a detailed answer.  

Recently, Drummond and Holte \cite{drummondholtekdd:00} have
demonstrated an intriguing dual to the \rocch.  Their ``cost curves''
represent expected costs explicitly, rather than as slopes of
iso-performance lines, and thereby provide an insightful alternative
perspective for visualization.

Note: An implementation of the \rocch\ method in Perl is publicly available.
The code and related papers may be found at:
\url{http://www.hpl.hp.com/personal/Tom_Fawcett/ROCCH/}.

\section{Acknowledgments}

Much of this work was done while the authors were employed at the Bell
Atlantic Science and Technology Center.  We thank the many with whom we have
discussed ROC analysis and classifier comparison, especially Rob Holte, George
John, Ron Kohavi, Ron Rymon, and Peter Turney.  We thank Andrew Bradley for
supplying data from his analysis.

\bibliographystyle{theapa}
\bibliography{final}
\end{document}